\documentclass[letterpaper, 10pt, conference]{ieeeconf}

\pdfoutput=1 

\IEEEoverridecommandlockouts
\usepackage{amsmath}
\usepackage{amsfonts}
\usepackage{amssymb}
\usepackage{bbold}  
\usepackage{multirow}

\usepackage{mdframed}

\usepackage[caption=false,font=footnotesize]{subfig}

\ifx\pdfoutput\undefined
\usepackage{graphicx} 
\else
\ifx\pdfoutput\relax
\usepackage{graphicx} 
\else
\ifnum\pdfoutput>0
\usepackage[pdftex]{graphicx} 
\else
\usepackage{graphicx} 
\fi
\fi
\fi

\usepackage{tabularx}
\usepackage{booktabs}

\usepackage{xspace} 

\usepackage{url}

\usepackage{doi}

\usepackage{flushend} 

\newcommand{\argmin}[1]{\ensuremath{\underset{#1}{\textrm{arg min}}\;}}
\newcommand{\argmax}[1]{\ensuremath{\underset{#1}{\textrm{arg max}}\;}}

\usepackage[textsize=footnotesize]{todonotes}

\makeatletter 
\let\NAT@parse\undefined
\makeatother

\usepackage[numbers,compress]{natbib}
\bibpunct{[}{]}{,}{n}{}{;}

\usepackage{hyperref}

\title{\LARGE \bf
Learning Articulated Motion Models from Visual and Lingual Signals
}

\author{%
  \authorblockN{Zhengyang Wu}%
  \authorblockA{%
    Georgia Tech\\%
    Atlanta, GA 30332\\%
    {\tt\small zwu66@cc.gatech.edu}}%
  \and
  \authorblockN{Mohit Bansal}%
  \authorblockA{TTI-Chicago\\%
    Chicago, IL 60637\\%
    {\tt\small mbansal@ttic.edu}}%
  \and
  \authorblockN{Matthew R.~Walter}%
  \authorblockA{TTI-Chicago\\%
    Chicago, IL 60637\\%
    {\tt\small mwalter@ttic.edu}}%
}

\begin{document}

\newpage

\maketitle

\begin{abstract}
    In order for robots to operate effectively in homes and
    workplaces, they must be able to manipulate the articulated
    objects common within environments built for and by humans. Previous
    work learns kinematic models that prescribe this manipulation from
    visual demonstrations. Lingual signals, such as natural language
    descriptions and instructions, offer a complementary means of
    conveying knowledge of such manipulation models and are suitable
    to a wide range of interactions (e.g., remote manipulation). In
    this paper, we present a multimodal learning framework that
    incorporates both visual and lingual information to estimate the
    structure and parameters that define kinematic models of
    articulated objects. The visual signal takes the form of an RGB-D
    image stream that opportunistically captures object motion in an
    unprepared scene. Accompanying natural language descriptions of
    the motion constitute the lingual signal.  We present a
    probabilistic language model that uses word embeddings to
    associate lingual verbs with their corresponding kinematic
    structures. By exploiting the complementary nature of the visual
    and lingual input, our method infers correct kinematic structures
    for various multiple-part objects on which the previous
    state-of-the-art, visual-only system fails.  We evaluate our multimodal
    learning framework on a dataset comprised of a variety of household
    objects, and demonstrate a $36\%$ improvement in model accuracy
    over the vision-only baseline. 
\end{abstract}

\section{Introduction} \label{sec:intro}

As robots move off factory floors and into our homes and workplaces,
they face the challenge of interacting with the articulated objects
frequently found in environments built by and for humans (e.g., drawers, ovens,
refrigerators, and faucets). Typically, this interaction is
predefined in the form of a manipulation policy that must be
(manually) specified for each object that the robot is expected to
interact with. In an effort to improve efficiency and
generalizability, recent work employs visual demonstrations to learn
representations that describe the motion of these parts in the form of
kinematic models that express the rotational, prismatic, and rigid
relationships between object
parts~\citep{sturm11,huang12,katz13,pillai14}. These structured
models, which constrain the manifold on which the object's motion
lies, allow for manipulation policies that are more efficient and
deliberate. However, such visual cues may be too time-consuming to
provide or may not be readily available, such as in the case of a
disaster relief scenario in which a user is remotely commanding a robot over a
bandwidth-limited channel. Further, reliance solely on vision 
makes these methods sensitive to common errors in object
segmentation and tracking that occur as a result of clutter,
occlusions, and a lack of
visual features. Consequently, most existing systems require scenes to be
free of distractors and that object parts be labeled with fiducial markers.

Lingual input in the form of natural language descriptions and
instructions offer a flexible, bandwidth-efficient medium that humans
can readily use to convey knowledge of an object's operation. Such
lingual descriptions of an articulated motion also provide a source of
information that is complementary to visual input. Thus, these
descriptions can be used to overcome some of the limitations of using
visual-only observations, e.g., by providing cues regarding the number
of parts that comprise the object or the motion type (e.g.,
rotational) between a pair of parts. In this work, we present a
multimodal learning framework that estimates the kinematic structure
and parameters of complex multi-part objects using both visual and
lingual input, and performs substantially better than visual-only
systems.

\begin{figure}
    \centering
    \includegraphics[width=0.45\textwidth]{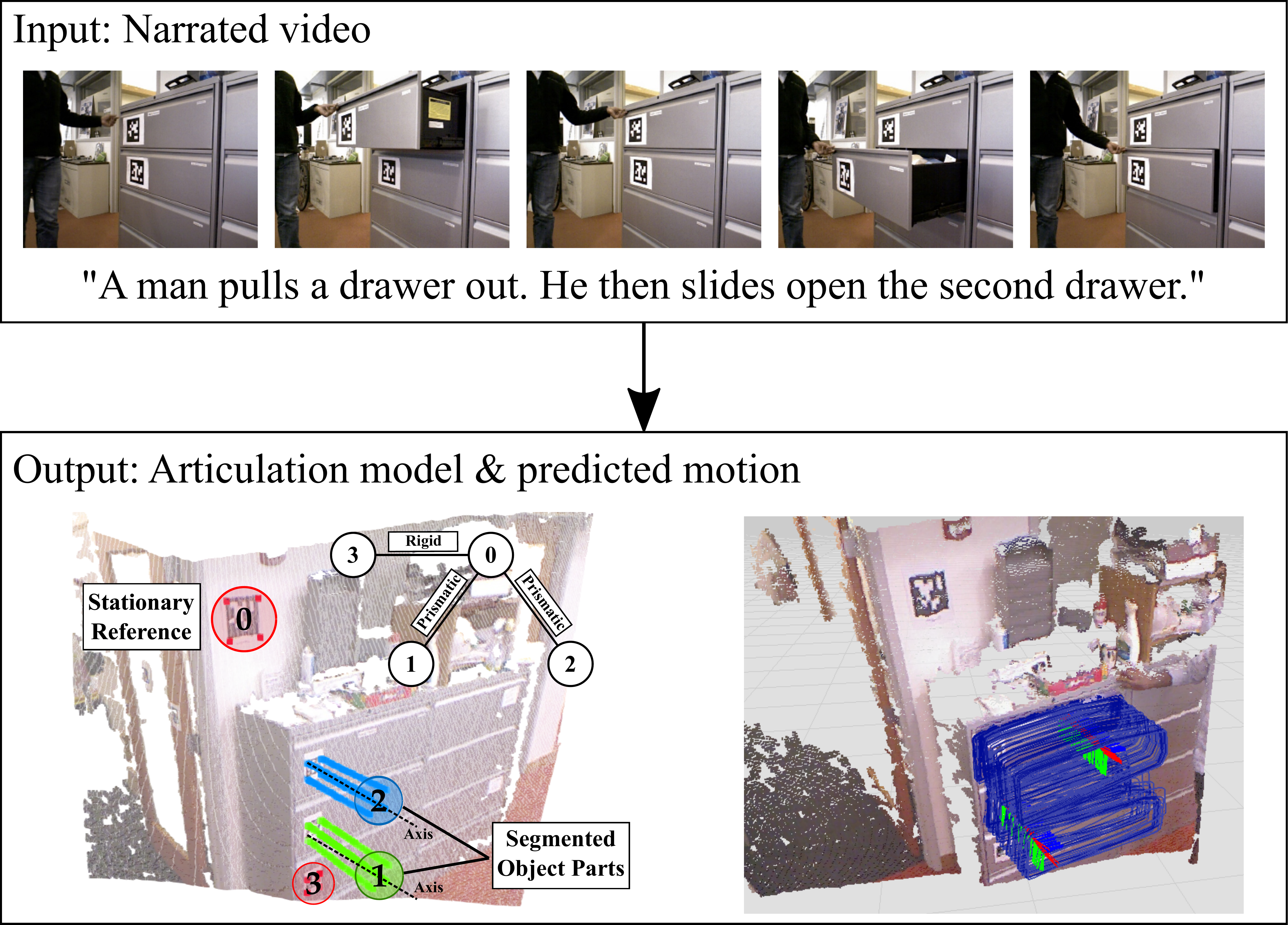}
    \caption{Our framework learns the kinematic model that governs the
      motion of articulated objects (lower-left) from narrated RGB-D
      videos. The method can then use this learned model to
      subsequently predict the motion of an object's parts (lower-right).} \label{fig:teaser}
\end{figure}
Our effort is inspired by the recent attention that has been paid to
the joint use of vision and language as complementary signals for
machine
perception~\citep{ordonez11, mitchell2012midge, sharma12,
  kong2014what, ramanathan2014linking, guadarrama14, karpathy-15,
  vinyals15, xu-15, kiros-14, mao-14, donahue-14, vinyals-14,
  chen-15,yao15, srivastava15, venugopalan15, mei16}. Much
of the work in multimodal learning considers the problems of image
caption generation and visual coreference resolution. Instead, we
leverage the joint advantages of these two modalities in order to
estimate the structure and parameters that define kinematic models of
complex, multi-part objects such as doors, desks, chairs, and
appliances.

Our multimodal learning framework first extracts noisy observations of
the object parts and their motion separately from the visual and
lingual signals. It then fuses these observations to learn a
probabilistic model over the kinematic structure and model parameters
that best explain the motion observed in the visual and lingual
streams. Integral to this process is an appropriate means of
representing the ambiguous nature of observations gleaned from natural
language descriptions. We propose two probability models that capture
this uncertainty based upon the similarity between the natural
language text and a representative reference word set (for each model
type) in a word embedding space. The first takes the form of a hard assignment of verbs in the
description to the nearest kinematic model type (e.g.,
rotational or prismatic) in the embedding space. The second takes the form of a soft assignment,
representing the likelihood of the
lingual observations in terms of the similarity between the input text
and the reference embeddings.

Our contributions include a multimodal approach to learning kinematic
models from visual and lingual signals, the exploration of different
language grounding methods to align action verbs and kinematic models,
and the examination of various language priors in our
learning framework. We evaluate our method on a dataset of video-text
pairs demonstrating the motion of common household objects, and
achieve notable improvement over the previous
state-of-the-art, which only uses visual information. The word
embedding-based soft and hard language models yield improvements of $21\%$
and $36\%$, respectively, demonstrating the promise of a multimodal
learning framework that exploits both visual and lingual information.


\section{Related Work} \label{sec:related}

Our goal is to enable robots to learn kinematic models with minimal supervision from human
demonstrations. This requires solutions that can mitigate the
complexity and clutter typical of human-occupied environments, without the
need for additional infrastructure (e.g., visual fiducials).

Recent work considers the problem of learning articulated models based
upon visual observations of demonstrated motion. Several methods
formulate this problem as bundle adjustment, using
structure-from-motion methods to first segment an articulated object
into its compositional parts and to then estimate the parameters of
the rotational and prismatic degrees-of-freedom that describe inter-part
motion~\citep{yan06,huang12}. These methods are prone to erroneous
estimates of the pose of the object's parts and of the inter-part
models as a result of outliers in visual feature
matching. Alternatively, \citet{katz10} propose an active learning framework
that allows a robot to interact with articulated objects to induce
motion. This method operates in a deterministic manner, first assuming
that each part-to-part motion is prismatic. Only when the residual
error exceeds a threshold does it consider the alternative rotational
model. Further, they estimate the models based upon interactive
observations acquired in a structured environment free of clutter,
with the object occupying a significant portion of the RGB-D sensor's
field-of-view. \citet{katz13} improve upon the complexity of this method
while preserving the accuracy of the inferred models. This method is
prone to over-fitting to the observed motion and may result in overly
complex models to match the observations. \citet{hausman15} similarly
enable a robot to interact with the object and describe a
probabilistic model that integrates observations of fiducials with
manipulator feedback. Meanwhile, \citet{sturm11} propose a
probabilistic approach that simultaneously reasons over the likelihood
of observations while accounting for the learned model
complexity. Their method requires that the number of parts that
compose the object be known in advance and that fiducials be placed on
each part to enable the visual observation of motion. More recently,
\citet{pillai14} propose an extension to this work that uses novel
vision-based motion segmentation and tracking that enables model
learning without prior knowledge of the number of parts or the
placement of fiducial markers. Our approach builds upon this method
with the addition of natural language descriptions of motion as an
additional observation mode in a multimodal learning framework.

Meanwhile, recent work in the natural language processing community has focused
on the role of language as a means of commanding~\citep{kollar-10,
  matuszek-10, tellex-11, chen-11, artzi-13, mei16} and sharing
spatial information~\citep{walter13, duvallet14, hemachandra-15} with
robots. We use language for the novel and more complex task of
learning object articulation in terms of kinematic motion
models. Meanwhile, other methods have similarly used visual and
lingual cues in a multimodal learning framework for such tasks as
image and video caption synthesis~\citep{ordonez11, mitchell2012midge,
  sharma12, guadarrama14, karpathy-15, vinyals15, xu-15, kiros-14,
  mao-14, donahue-14, vinyals-14, chen-15, yao15, srivastava15,
  venugopalan15}, visual coreference resolution~\citep{kong2014what,
  ramanathan2014linking}, visual question-answering~\cite{antol-15},
and understanding cooking videos paired with
recipes~\cite{malmaud15}. Our work shares similar goals, particularly
in the context of action inference based on joint visual-lingual cues.


\section{Multimodal Learning Framework} \label{sec:approach}

\begin{figure*}
  \includegraphics[width=\textwidth]{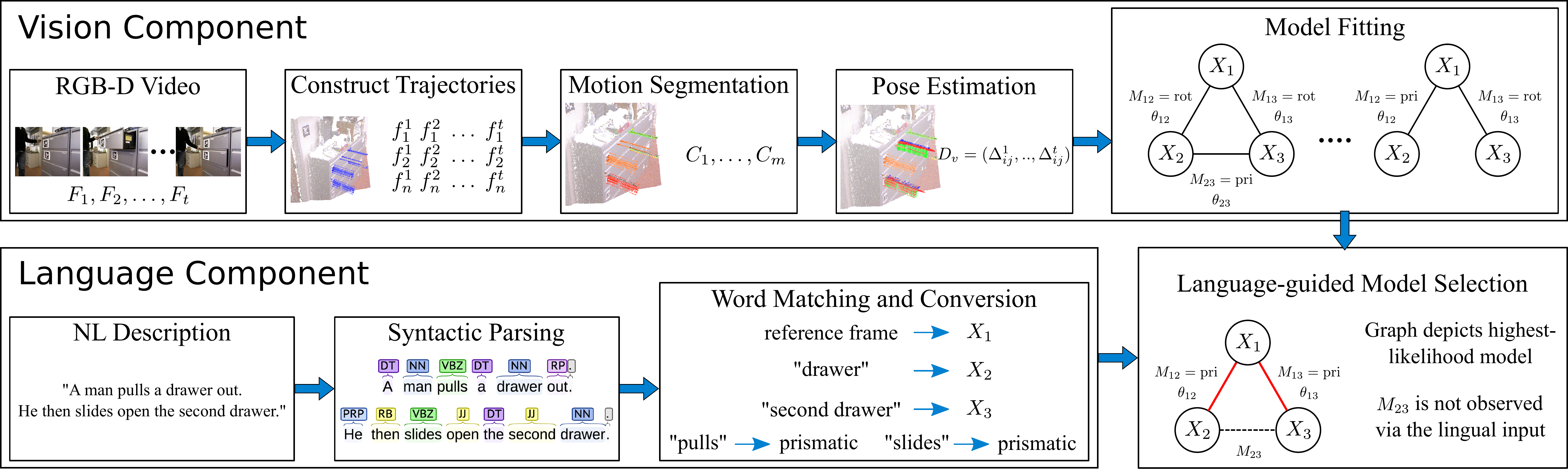}
  \caption{Our multimodal articulation learning framework first
    identifies clusters of visual features that correspond to
    individual object parts. It then uses these feature trajectories
    to estimate the model
    parameters, assuming an initial estimate of the kinematic type
    associated with each edge in
    the graph. The method then uses natural language
    descriptions of the motion to estimate the kinematic type of each edge through a
    probabilistic language model.}
  \label{fig:architecture}
\end{figure*}
Given an RGB-D video paired with the corresponding natural language
description (alternatively, an instruction or caption) of an
articulated object's motion, our goal is to infer the structure and
parameters of the object's kinematic model. Adopting the formulation
proposed by~\citet{sturm11}, we represent this model as a
graph, where each vertex denotes a different part of the object (or the
stationary background) and edges denote the existence of constrained
motion (e.g., a linkage) between two parts (Fig.~\ref{fig:teaser}). More formally, we estimate
a \emph{kinematic graph} $G = (V_G, E_G)$ that consists of vertices $V_G$ for each object part and edges
$E_G \subset V_G \times V_G$ between parts whose relative motion is
kinematically constrained. Associated with each edge $(ij) \in E_G$ is its kinematic type
$M_{ij} \in \{\textrm{rotational},
\textrm{prismatic},\textrm{rigid}\}$
as well as the corresponding parameters $\theta_{ij}$, such as the
axis of rotation and the range of motion (see
Fig.~\ref{fig:architecture}, lower-right). We take as
input visual $D_v$ and lingual $D_l$ observations of the type and
parameters of the edges in the graph. Our method then uses this
vision-language observation pair $D_z = \{D_v,D_l\}$ to infer the maximum a posteriori kinematic structure and model
parameters that constitute the kinematic graph:
\begin{subequations}
    \begin{align}
      \hat{G} &= \argmax{G} p(G \vert D_z) \\
              &= \argmax{G} p(\{M_{ij}, \theta_{ij} \vert (ij) \in E_G\} \vert D_z)\\
              &= \argmax{G} \prod_{(ij)\in E_G}p(M_{ij}, \theta_{ij}
                \vert D_z)
    \end{align}
\end{subequations}

Due to the complexity of joint inference, we adopt the procedure
described by \citet{sturm11} and use a two-step inference procedure
that alternates between model parameter fitting and model structure
selection steps (Fig.~\ref{fig:architecture}). In the first step, we assume
a particular kinematic model type between each object $i$ and $j$
(e.g., prismatic), and then estimate the kinematic parameters based on
the vision data (relative transformation between the two objects) and
the assumed model type $M_{ij}$. We make one such assumption for each
possible model type for each object pair.

In the model selection step, we then use the natural language
description to infer the kinematic graph structure that best expresses
the observation. While our previous work~\citep{pillai14} provides
visual observations of motion without the need for fiducials, it
relies upon feature tracking and segmentation that can fail when the
object parts lack texture (e.g., metal door handles) or when the scene
is cluttered. Our system incorporates language as an additional,
simpler, complementary observation of the motion, in order to improve
the robustness and accuracy of model selection.

\subsection{Vision-guided Model Fitting} \label{sec:vision-model}

Given the RGB-D video of the motion, we employ the vision pipeline of
\citet{pillai14} to arrive at a visual observation of the trajectory
of each object part (Fig.~\ref{fig:architecture}). The method first identifies a set of 3D feature
trajectories that correspond to different elements in the scene,
including the object parts, background, and clutter (the distinction of
which is not known a priori). These trajectories are then grouped to
form rigid clusters according to the similarity of their relative
motion in an effort to associate a cluster to each object part as well
as to the background. Next, the method estimates the 6-DOF pose
trajectory of each cluster (object part). We refer the reader to
\citet{pillai14} for specific details regarding the visual pipeline.

The 6-DOF pose trajectories constitute the visual observation of the
motion $D_v$. Our framework uses these trajectories to estimate the
parameters of a candidate kinematic model during the model fitting
step. Specifically, we find the kinematic parameters that best
explain the visual data given the assumed model
\begin{equation} \label{eqn:model-fitting}
    \hat{\theta}_{ij} = \argmax{\theta_{ij}} p(D_v \vert \hat{M}_{ij}, \theta_{ij}),
\end{equation}
where
$D_v = (\Delta_{ij}^1, ..., \Delta_{ij}^t), \forall (ij) \in E_G $ is
the sequence of observed relative transformations between the poses of
two object parts $i$ and $j$, and $\hat{M}_{ij}$ is the current estimate of their
model type. We perform this optimization over the joint kinematic
structure defined by the edges in the graph~\citep{sturm11}.

\subsection{Language-guided Model Selection}

Methods that solely use visual input are sensitive to the effects of
scene clutter and the lack of texture, which can result in erroneous
estimates for the structure and parameters of the kinematic
model~\citep{pillai14}. We incorporate lingual observations into our
framework in order to reduce errors that result from the failure in
the visual pipeline, and also to add complementary observational
information.

Specifically, we consider a natural language caption $D_l$ that
describes the motion observed in the video. Given this description, we
infer the maximum a posteriori model type for each pair of object
parts according to the caption. We require that the natural language
descriptions adhere to
a known grammar and that it narrate at least one
motion present in the video\footnote{As determined by the number of verbs.}
(otherwise, our method estimates the kinematic graph based solely on
the visual observation). Note that we do not assume that valid
captions provide an unambiguous description of the motion, but rather
consider a distribution over the lingual observation, which provides
robustness to ``noisy'' captions. We employ
the following procedure (bottom-left of Fig.~\ref{fig:architecture}) to
convert a natural language description into a structured caption
representation:
\begin{enumerate}
\item Perform word tokenization and part-of-speech tagging of the natural language description to obtain object nouns and action verbs.
\item Align the object nouns to the visual motion trajectories.
\item Classify the action verbs into kinematic model types (i.e., ``prismatic,'' ``rotational,'' or ``rigid'').
\end{enumerate}
Next, we discuss each of these steps in detail.

\subsubsection{Preprocessing Natural Language}

Our system first extracts object- and motion-relevant cues from the
natural language caption in the form of nouns that denote object parts
and verbs that describe their motion. Nouns that refer to the agent
(e.g., ``man'' or ``person'') are ignored. We use the Stanford CoreNLP
pipeline, tokenizer, and POS-tagger to identify the nouns and verbs
 in the caption~\cite{manning14}.

\subsubsection{Matching Object Nouns with Trajectories}

Given the set of nouns in the narration, we next seek to identify the
corresponding object parts in the visual clusters. We enumerate the
space of possible noun-cluster correspondences and choose the
noun-cluster assignment with the lowest error, which we define shortly
(Eqn.~\ref{eqn:bic-error}). In practice, this exhaustive search is not
a bottleneck as most objects that we are interested in, including
those found in the home, contain a manageable number of parts. Note
that we also investigated the use of vision-based object recognition
to reduce this search space~\citep{girschick14}, but found the
recognition accuracy to be insufficient for such
tasks (detectors were prone to false negatives and tend to predict
holistic object classes like ``bicycle'' instead of their parts like
``bicycle wheel'' and ``bicycle frame,'' which is necessary for our
task).

\subsubsection{Convert Action Verbs to Kinematic Model
  Type} \label{sec:convert_action}
\begin{figure}[!t]
    \centering
    \begin{mdframed}[align=center]{\footnotesize \textbf{Prismatic}:
          \emph{pull}, \emph{push}, shift, \emph{move},
          \emph{close}, remove, tug, yank, dislocate, extract,
          jerk, thrust, poke, prod, shove, displace, stretch, squeeze,
          fasten, draw, join, insert, embed, enter, exit, implant,
          inject, introduce, stick, admit, infuse, inlay,
          instill, place, set, penetrate, withdraw, intrude, slide.}\\[10pt]
          {\footnotesize \textbf{Rotational}: bend, yaw, turn, spin, whirl,
          \emph{move}, \emph{pull}, \emph{push}, \emph{close},
          revolve, rotate, gyre, gyrate, pivot, swivel, twist, twirl,
          circle, roll, reel, wheel, round, wrench, screw, tighten,
          swing, cycle, bow, flex, wind, spiral, twine, loosen.}
    \end{mdframed}
    \caption{Our manual dictionary of motion verbs for prismatic and
      rotational kinematic types. Words in italics are shared between
      the two dictionaries.} \label{fig:dictionary}
\end{figure}

The final step of our framework converts the parsed action verbs to
either ``prismatic'' or ``rotational'' kinematic model
types.\footnote{We currently assume no action verbs for the ``rigid''
  type and default to the visual observation.} A simple means of
performing this conversion, which we treat as an oracle, is to
manually create verb dictionaries that span the variety of words that
can be used for each of the rotational and prismatic motion
types. Figure~\ref{fig:dictionary} enumerates the words that define
our dictionary. Note that some words are shared by both dictionaries
(e.g., ``push'' can be used to describe both prismatic and rotational
motion), in which case the lingual observation would have equal
likelihood for different models.

The manual dictionary simply provides an oracle baseline. Our system employs a
general, non-manual approach to convert verbs to their corresponding
type. Specifically, we embed words in a learned, high-dimensional
space and use their relative distances in this space to identify model
correspondence. First, we select a small seed dictionary
\mbox{$W = {w_1,w_2,...,w_s}$} that includes the $s$ most common words
for each model type (we use $s=3$ in our experiments) from the full
manual dictionary in Figure~\ref{fig:dictionary}. These can be thought
of as the seed clusters representing each model type. A seed
dictionary is important to construct the model type's centroid vector
because there is no single canonical word that can represent the
entire meaning of a general kinematic model type such as
``prismatic.'' We use the set
$W_\textrm{prismatic}=\{\textrm{``shift'', } \textrm{``insert'', } \textrm{``extract''}\}$
as the seed cluster for the prismatic model and the set
$W_\textrm{rotational}= \{\textrm{``rotate'', }
\textrm{``circle'', } \textrm{``twist''}\}$ as the seed cluster for the rotational model
in our experiments.  Next, we convert each word to a
$d$-dimensional word embedding space using word2vec~\cite{mikolov13},
a popular neural language model. We compute the mean over the word vectors in each
seed dictionary to arrive at a ``centroid'' vector
$\vec{w}_\textrm{prismatic}$ and $\vec{w}_\textrm{rotational}$  that represents the
corresponding kinematic model type. Given a new unseen verb $w_\textrm{new}$
from a test sentence, we project it to the same word embedding space
(using word2vec) and then compare it with with the centroid vector of
each model type according to cosine distance. The model type
$M_\textrm{new}$ with the smallest distance is set as the model type
of this action verb embedding $\vec{w}_\textrm{new}$:
\begin{equation}
    M_\textrm{new} = \argmin {m\in\{\textrm{rot},\textrm{pri}\}}
    \textrm{dist}(\vec{w}_m, \vec{w}_\textrm{new}).\label{eqn:nearest_cluster}
\end{equation}
We require this distance to be lower than that of the other model
by a margin (we use $0.1$ in our experiments). Otherwise, we treat
the word as ambiguous and assign it to both kinematic models.

\subsection{Combining Visual and Lingual Observations}

The final step in our framework selects the kinematic graph structure
$\hat{\mathcal{M}} = \{\hat{M}_{ij}, \forall (ij) \in E_G \}$ that best explains the visual and lingual observations
$D_z = \{D_v, D_l\}$ from the space of all possible kinematic
graphs. We do so by maximizing the conditional posterior over the
model type associated with each edge in the graph $(ij) \in E_G$:
\begin{subequations}
    \begin{align}
      \hat{M}_{ij} &= \argmax{M_{ij}} p(M_{ij} \vert D_z) \\
      &= \argmax{M_{ij}} \int p(M_{ij},\theta_{ij} \vert D_z)  d\theta_{ij}
    \end{align}
\end{subequations}
Evaluating this likelihood is computationally prohibitive, so we use
the Bayesian Information Criterion (BIC) score as an approximation
\begin{equation}
    BIC(M_{ij}) = -2 \log p(D_z \vert M_{ij}, \hat{\theta}_{ij}) + k \log n,
\end{equation}
where $\hat{\theta}_{ij}$ is the maximum likelihood parameter estimate
(Eqn.~\ref{eqn:model-fitting}), $k$ is the number of parameters of the current model and $n$ is
the number of visual and lingual observations. We choose the model with the
lowest BIC score:
\begin{equation} \label{eqn:bic-error}
    \hat{M}_{ij} = \argmin{M_{ij}} BIC(M_{ij})
\end{equation}

While our previous method~\citep{pillai14} only considers visual
observations, our new framework performs this optimization over the joint
space of visual and lingual observations. Consequently, the BIC score
becomes
\begin{equation}
    \begin{split}
        BIC(M_{ij}) = -2 \Bigl(&\log p(D_v \vert M_{ij},
        \hat{\theta}_{ij})\\
        &+ \log p(D_l \vert M_{ij}, \hat{\theta}_{ij})\Bigr) + k\log n,
    \end{split}
\end{equation}
where we have made the assumption that the lingual and visual
observations are conditionally independent given the model and
parameter estimate. Here, the language model can take one of two
forms. The first acts as a hard assignment of a verb to its
corresponding model type, whereby we assign a likelihood
of one to the model whose centroid vector is closest
 in the embedding space (Eqn.~\ref{eqn:nearest_cluster}) and zero to the other, subject to a
margin. For ambiguous words, i.e., those that can be associated with
either model type according to the margin, the probability is
equal ($0.5$) for both of the candidate kinematic models. The second
form acts as a soft assignment, setting the model conditional distributions
$p(D_l \vert M_{ij}, \hat{\theta}_{ij})$ according to the cosine
similarity between the word in the input associated with the motion
$\vec{w}_\textrm{verb} \in D_l$ and the model's centroid vector
$\vec{w}_m \in \{\vec{w}_\textrm{prismatic}, \vec{w}_\textrm{rotational}\}$ in the
embedding space
\begin{equation}
    p(D_l \vert M_{ij}, \theta_{ij}) = \textrm{dist}(\vec{w}_m, \vec{w}_\textrm{verb}).
\end{equation}

We then estimate the overall kinematic structure by solving for the
minimum spanning tree of the graph, where we define the cost of each
edge as
$\textrm{cost}_{ij} = -\log p(M_{ij}, \theta_{ij} \vert
D_z)$.
Such a spanning tree constitutes the kinematic graph that best
describes the visual and lingual observations.


\section{Results} \label{sec:results} 
\begin{table*}[!th]
    \centering
    \caption{Overall Performance of Our Framework \label{tab:main}}
    \def\arraystretch{1.3}
    \begin{tabular}{@{}cccccccc@{\hspace{2pt}}ccc@{}}
      \toprule
      &&&&& \multicolumn{2}{c}{Vision-Only} && \multicolumn{3}{c}{Our Framework}\\
      \cmidrule{6-7} \cmidrule{9-11} 
      & Object & $N^*$ & $N_v$ & $S_v$ & $S_{h}$ & $S_{s}$ && $S_{h}$ & $S_{s}$ & $e_\textrm{param}$\\
      \midrule
      \multirow{4}{*}{Single-Part} & Door & 1  & 1,1  & 2/2 & 2/2  & 2/2 && 2/2 & 2/2 & \hphantom{0}1.86$^\circ$\\
      %
      & Chair & 1 & 1,1,1 & 3/3 & 2/3 & 2/3 && \textbf{3/3}  & \textbf{3/3} & \hphantom{0}3.34$^\circ$\\
      %
      & Refrigerator & 1 & 1,1,1,1 & 4/4 & 3/4 & 3/4 && {\bf 4/4} & {\bf 4/4} & \hphantom{0}5.74$^\circ$\\
      %
      & Microwave & 1 & 1,1 & 2/2 & 2/2 & 2/2 && 2/2 & 2/2 & \hphantom{0}2.02$^\circ$\\
      \midrule
      %
      \multirow{4}{*}{Multi-Part} & Drawer & 2 & 1,2 & 1/2 & 0/2 & 1/2 && \textbf{1/2} & \textbf{2/2} & \hphantom{0}0.11$^\circ$\\
      %
      & Monitor & 3 & 1,1,1,1,3,3 & 2/6 & 1/6 & 4/6 && {\bf 2/6} & {\bf 6/6} & \hphantom{0}7.27$^\circ$\\
      %
      & Bicycle & 3 & 1,2,2,2,2,2 & 0/6& 0/6 & 3/6 && 0/6 & \textbf{6/6} & 11.33$^\circ$\\
      %
      & Chair & 2 & 1,2,2 & 2/3 & 0/3 & 1/3 && \textbf{2/3} & \textbf{3/3} & \hphantom{0}3.05$^\circ$\\
      \bottomrule
      \end{tabular}
\end{table*}
We evaluate our framework on $28$ RGB-D videos in which a user
manipulates a variety of common
household and office objects (e.g., a microwave, refrigerator, and
drawer). AprilTags~\cite{olson11} were
placed on each of the objects parts and used as an observation of
ground-truth motion. We mask the AprilTags when running the visual
pipeline so as to not affect feature extraction. Of the $28$ videos,
$13$ involve single-part objects and $15$ involve multi-part
objects. The single-part object videos are used to demonstrate that
the addition of lingual observations can only improve the accuracy of
the learned kinematic models. The extent of these improvements on
single-part objects is limited by the relative ease of inference of
single degree-of-freedom motion. In the case of multi-part objects,
the larger space of candidate kinematic graphs makes visual-only
inference challenging as feature tracking errors may result in
erroneous estimates of the graph structure. These experiments are meant to evaluate the
extent to which multimodal learning improves model selection.

After watching the videos, we asked a user to provide a single caption
for each video. Before doing so, we provided the user with some examples of potential
captions that discuss the movement of the individual parts as
opposed to single, high-level captions. An example of such a narration
is ``A man pushes the bicycle frame forward. The front wheel is
spinning. The back wheel is rotating.''  as opposed to the high-level
caption ``A man pushes a bicycle forward,'' which would not be
sufficient (because our system is unable to associate ``bicycle'' with
only the frame). This is similar to discussions for image and video
captioning and question-answering research, where it is well-known
that a more detailed, database-like caption is more useful for
capturing multiple salient events in the image/video, and for
answering questions made of them~\citep{antol-15}.

\subsection{Evaluation Metrics and Baselines}

We estimate the ground-truth kinematic models by performing MAP
inference based upon the motion trajectories observed using
AprilTags. We denote the resulting kinematic graph as $G^*$. The
kinematic type and parameters for each object part pair are denoted as
$M^*_{ij}$ and $\theta^*_{ij}$, respectively. Let $\hat{G}$,
$\hat{M}_{ij}$, $\hat{\theta}_{ij} $ be the estimated kinematic
graph, kinematic type, and parameters for each object pair from the
RGB-D video, respectively.

The first metric that we consider evaluates whether the vision
component estimates the correct number of parts.  We determine the
ground-truth number of parts as the number of AprilTags observed in
each video, which we denote as $N^*$. We indicate the number of parts
(motion clusters) identified by the visual pipeline as $N_v$. We
report the average success rate when using only visual observations as
\mbox{$S_v = \frac{1}{K} \sum_{k=1}^{K} \mathbb{1}(N_v^k = N^{k*})$}, where
$K$ is the number of videos for each object type.

Next, we consider two metrics that assess the ability of each method
to estimate a graph with the same kinematic model as the ground truth
$G^*$. The first metric requires that the two graphs have the same
structure, i.e.,
$\hat{M}_{ij} = M^*_{ij}, \forall (ij) \in E_{\hat{G}} =
E_{G^*}$.
This equivalence requires that vision-only inference yields the
correct number of object parts and that the model selection framework
selects the correct kinematic edge type for each pair of object
parts. We report this ``hard'' success rate $S_h$ in terms of the
fraction of demonstrations for which the model estimate agrees with
ground truth. Note that this is bounded from above by fraction for
which the vision component estimates the correct number of parts. The second ``soft''
success rate (denoted by $S_{s}$) employs a relaxed
requirement whereby we only consider the inter-part relationships
identified from vision, i.e.,
$\hat{M}_{ij} = M^*_{ij}, \forall (ij) \in E_{\hat{G}} \subset
E_{G^*}$.
In this way, we consider scenarios for which the visual system detects
fewer parts than are in the ground-truth model. In our experiments, we
found that $\hat{G}$ is a sub-graph of $G^*$, so we only require that
the model type of the edges in this sub-graph agree between both
graphs. The metric reports the fraction of total demonstrations for
which the estimated kinematic graph is a correct sub-graph of the ground-truth
kinematic graph.

Once we have the same kinematic models for both $\hat{G}$ and $G^*$,
we can compare the kinematic parameters $\hat{\theta}_{ij}$ to the
ground-truth values $\theta^*_{ij} $ for each inter-part model
$\hat{M}_{ij}$. Note that for the soft metric, we only compare
kinematic parameters for edges in the sub-graph, i.e.,
\mbox{$\forall (ij) \in E_{\hat{G}} \subset E_{G^*}$}. We define the parameter
estimation error for a particular part pair as the angle between the two kinematic parameter axes
\begin{equation}
    \textrm{e}_{ij} = \arccos \frac{\hat{\theta}_{ij} \cdot \theta^*_{ij}}{\lVert \hat{\theta}_{ij} \rVert \lVert \theta^*_{ij} \rVert},
\end{equation} 
where we use the directional and rotational axes for prismatic and
rotational degrees-of-freedom, respectively. We measure the overall parameter estimation
error $e_\textrm{param}$ for an object as the average parameter estimation error over each
edge in the object's kinematic graph. We report this error further
averaged over the number of demonstrations.

\subsection{Results and Analysis}

Table~\ref{tab:main} summarizes the performance of our multimodal
learning method using our embedding-based language model with hard alignment, comparing
against the performance of the vision-only
baseline~\citep{pillai14}. The table indicates the ground-truth number
of parts for each object ($N^*$), a list of the number of parts identified using
visual trajectory clustering for each demonstration ($N_v$), and the
fraction of videos for which the correct number of parts was
identified ($S_v$). We then present the hard ($S_h$) and soft ($S_s$)
model selection rates for our method as well as for the baseline. Our
method bests the vision-only baseline in estimating the full kinematic
graph for five of the eight objects, matching its performance on the
remaining three objects. Specifically, our framework yields accurate
estimates of the full kinematic graphs for six more demonstrations
than the vision-only baseline, two more for single-part objects and
four more for multi-part objects, corresponding to a $21\%$ absolute improvement. Similarly, we are able to estimate
a valid sub-graph of the ground-truth kinematic graph for all 28
demonstrations, whereas the vision-only baseline fails to estimate
valid sub-graphs for ten of the videos (two for single-part
and eight for multi-part objects), corresponding to a $36\%$ absolute
improvement.\footnote{The soft alignment model results in improvements
of $14\%$ for the complete graph estimates and $21\%$ for the
sub-graphs estimates.} One notable object on which both
methods have difficulty is the bicycle for which the trajectory
clustering method was unable to identify the presence of the third
part (the wheel)
due to the sparsity of visual features on the wheels. Consequently,
neither method estimated the full kinematic graph for any video,
however our framework was able to exploit lingual cues to yield
accurate sub-graph estimates for each video. Similarly, clustering
failed to identify the three parts that comprise the monitor in all
but two videos, for which our method then estimated the
correct kinematic graph (and an accurate sub-graph for the remaining
four videos).

We then evaluate the accuracy of the parameters estimated by our
method by reporting the parameter estimation error for each object,
averaged over the set of videos. Note that it is difficult to compare
against the error of the vision-only baseline since it does not yield
accurate kinematic graphs for several of the videos. When the kinematic graph
estimates agree, however, the parameter estimation errors are
identical for the two methods, since they both estimate the parameters
from the visual data (Eqn.~\ref{eqn:model-fitting}).



%
%
\begin{table}[!htb]
    \centering
    \caption{Detailed success and failure analysis \label{tab:additional}}
    \def\arraystretch{1.3}
    \begin{tabular}{@{}clccccc@{}}
      \toprule
      & & \multicolumn{2}{c}{Manual Dict.} & \multicolumn{3}{c}{Word Embedding}\\
      \cmidrule{3-7}
      & & Success & Ambig. & Success & Ambig. & WA\\
      \midrule
      \multirow{4}{*}{\begin{tabular}[c]{@{}c@{}}Single-\\ Part\end{tabular}} & Door& 10/10 & 0/10 & 10/10 & 0/10 & 0/10\\
      & Chair      & 12/15 & 3/15 & 10/15 & 3/15 & 2/15\\
      & Fridge     & 17/20 & 3/20 & 17/20 & 3/20 & 0/20\\
      & Microwave & 10/10 & 0/10 & 10/10 & 0/10 & 0/10\\ \hline
      \multirow{4}{*}{\begin{tabular}[c]{@{}c@{}}Multi-\\ Part\end{tabular}} & Drawer & 10/10 & 0/10 & 10/10 & 0/10 & 2/10\\
      & Monitor    & 29/30 & 1/30 & 29/30 & 1/30 & 0/30\\
      & Bicycle    & 27/30 & 3/30 & 19/30 & 9/30 & 5/30\\
      & Chair      & 10/15 & 5/15 & 10/15 & 5/15 & 0/15\\
      \bottomrule
    \end{tabular}
\end{table}
In order to better understand the effects of variability in the linguistic
input, we then asked the user to generate four additional diverse captions for
each video. Table~\ref{tab:additional} presents the overall
performance on the complete set of five captions per
video\footnote{We consider a scenario to be successful iff our method
identifies the correct kinematic graph for the sub-graph consisting of object parts identified
via visual clustering.} when using our word embedding language model
with soft alignment. For comparison, we consider the result with
oracle alignment, whereby verbs are matched with their corresponding
kinematic type according to the manual dictionary
(Fig.\ref{fig:dictionary}). As can be expected, the model selection accuracy is greater for
some captions when using the oracle dictionary. We attribute this
difference to two primary factors. First, some
captions describe the motion of parts using words that are ambiguous in
their meaning. For example, several captions include the term
``pull,'' which may refer to both prismatic or rotational motion
according to both the manual dictionary and the word embedding
representation, i.e.,
$\textrm{dist}(\text{``pull''},W_\textrm{pri} ) \simeq
\textrm{dist}(\text{``pull''}, W_\textrm{rot})$,
where $ W_\textrm{pri}, W_\textrm{rot}$ are the vectors that represent
the two kinematic types. Second, the word embedding-based method may yield
inaccurate estimates of word similarity as a result of having been trained
on general-domain text. For example, while ``slide'' is only in the
manually defined dictionary for prismatic motion, the word embeddings
suggest that it is equidistant from the centroid for each of these types, i.e.,
$\text{dist}(\text{``slide''},W_\textrm{pri} ) \simeq
\textrm{dist}(\text{``slide''}, W_\textrm{rot})$.
Note that ambiguities that result from similarity in the word
embedding space distances
are different from ambiguities inherent in the verb itself.  We
attribute the former to the failure of general-domain word embeddings
and report this fraction in the last column of Table~\ref{tab:additional},
denoted as ``WA.'' The fraction that fail due to the ambiguity
inherent in the specific verb itself is denoted as ``Ambig.'' and the
fraction that are successful is represented as ``Success.''  The total
number of description-video pairs is calculated based up upon five
descriptions per video. Note that the multimodal nature of our model
allows the visual signal to mitigate ambiguity in the lingual
observation. In this way, it is possible to use visual cues to
overcome failures of the linguistic models just as we use the lingual
signal to mitigate failure of the visual pipeline.


\section{Conclusion} \label{sec:conclusion}

We have described a method that uses a joint combination of visual and
lingual signals to learn accurate probabilistic models that define the
structure and parameters of articulated objects. Our framework treats
linguistic descriptions of a demonstrated motion as a complementary
observation of the structure of kinematic linkages. We evaluate our
framework on a series of RGB-D videos paired with captions of common
household and office objects, and demonstrate that the use of lingual
cues results in improved model accuracy. Future work includes the
incorporation of vision-based object part recognition, using the
captions to mitigate noise in the visual recognition. We are also
exploring a word embedding representation better suited to this
specific domain as means of more efficiently using visual and lingual
signals for complex objects. Additionally, we are investigating
extending our model to the problem of predicting kinematic models of
novel objects, using natural language captions as a means of
transferring knowledge from known classes.


\bibliographystyle{IEEEtranN}
\bibliography{references}

\end{document}